\documentclass{article}

\usepackage[preprint]{corl_2026} 
\usepackage{xspace}
\usepackage{xcolor}
\usepackage{amsmath,amssymb,amsthm}
\usepackage{bm}
\usepackage{cleveref}
\usepackage{color}
\usepackage{cleveref}
\usepackage{algorithm}
\usepackage{algpseudocode}
\usepackage{wrapfig}
\usepackage{booktabs}
\usepackage{graphicx}
\usepackage{subcaption}
\usepackage{tabularx}
\usepackage{enumitem}
\usepackage{upgreek}

\newcommand{\method}{{SARL}\xspace}
\newcommand{\methodspelledout}{{Semantic Action Reinforcement Learning}\xspace}

\newcommand{\loose}{\looseness=-1}

\title{Adapting Generalist Robot Policies with\\Semantic Reinforcement Learning}

\author{
  Jagdeep Singh Bhatia, Andrew Wagenmaker, William Chen, Sergey Levine\\
  U.C. Berkeley\\
  \href{https://semantic-action-rl.github.io}{\color{orange}\texttt{semantic-action-rl.github.io}}
}

\begin{document}
\newgeometry{margin=1.25in} 
\maketitle


\begin{abstract}
Generalist robot policies learn a diverse repertoire of behaviors from large-scale pretraining. In principle, this makes them excellent priors for downstream adaptation via reinforcement learning (RL). In practice, however, standard RL methods leveraging this prior optimize directly over robot actions, requiring the base policy's action distribution to be close to that of a performant policy from the start. This assumption breaks down for complex or long-horizon tasks that fall outside the pretraining distribution.
Our key insight is that, for sufficiently expressive generalist policies, language prompts are an effective alternative space for learning to solve such tasks: modulating language inputs elicits skills already within the policy's repertoire, which can be composed to solve tasks beyond its zero-shot capabilities. We propose \methodspelledout (\method), which learns to optimize this prompt space through online interaction, treating the generalist policy as a controllable skill prior. 
Importantly, leveraging pretrained skills rather than learning new ones from scratch yields structured, semantically meaningful exploration and highly efficient online improvement, and learning to modulate prompts through experience grounds them in induced real-world behaviors for robust task-solving.
Across real-world settings and simulated benchmarks, we show \method unlocks fundamentally new capabilities---adapting VLA behavior to solve complex, long-horizon tasks---and significantly outperforms existing approaches for improving robot behavior in deployment.
\end{abstract}


\section{Introduction}

Vision-language-action (VLA) models can learn broad skill repertoires from pretraining~\citep{lbm, black2024pi0, kim2024openvla, intelligence2025pi_}, providing powerful priors over plausible robot behaviors. However, leveraging VLAs to solve new tasks remains challenging. While we could directly prompt models to carry out new tasks of interest, this strategy requires models to not only have the necessary skills in their repertoire, but to also deploy them correctly. This is especially a challenge for complex and long-horizon tasks --- models must not only break down the semantics of such long-horizon tasks into executable, atomic behaviors, but also \emph{ground} these behaviors in skills the robot can perform successfully. As such, directly prompting VLAs to carry out novel complex tasks typically fails. How can we better leverage the prior knowledge in general-purpose robot policies to learn to solve these tasks effectively? 

Our key insight is that learning to modulate a VLA's prompts through reinforcement learning (RL) is an effective way to probe VLA priors to efficiently learn to solve new tasks. In particular, modulating language inputs elicits skills already within a VLA's repertoire, yielding constrained, high-quality exploration. By grounding each prompt in the behavior it induces through real-world interaction, these skills can be composed to solve tasks beyond a VLA's zero-shot capabilities. Unlike prior robotic RL approaches that operate at the \emph{robot action} level, we propose learning at the \emph{semantic} level, enabling a more effective use of the behavioral priors encoded in VLAs. For example, to train an omelet-making robot, we can simply learn semantic language commands to steer the VLA: ``turn on the stove", ``open the fridge", ``grab the egg", and so on, rather than needing to learn the precise low-level robot actions required to execute each of these skills. Reusing pre-trained skills rather than discovering new ones from scratch presents a potentially much easier learning problem, since the search space can be naturally constrained to task-appropriate behaviors using the semantics of the task, but still allows for improvement over the base policy. For VLAs with sufficiently broad language-inducible skill sets, language-based steering produces exactly the kind of exploration RL demands: highly expressive yet always grounded in reasonable actions, enabling efficient learning.

Our proposed approach, \methodspelledout (\method), instantiates this capability through an online RL loop that \emph{treats the VLA's language instructions as an action space}. \method uses interactions with the world to learn which language prompts will actually lead the VLA to complete the task of interest. By treating the VLA's language prompt as an ``action,'' we can lift the learning problem from low-level robot action space to semantic language space. By grounding prompts in real-world experience, we can efficiently refine which of them actually lead to success.

Our core contribution is an RL method that learns over the \emph{language inputs} into a VLA, enabling substantially more structured exploration and faster learning. Our results---both in simulation and on real-world robots---highlight that \method unlocks fundamentally new capabilities, enabling us to efficiently learn to solve complex, long-horizon tasks that are infeasible for 1) existing RL approaches, 2) zero-shot VLA deployment, and 3) VLAs deployed with high-level language steering (e.g., from a VLM with in-context learning). Finally, we investigate why \method learns effectively, and find that it is not only because \method decomposes complex goals into achievable skills, but that \method also learns a grounded mapping between semantic actions and the physical behaviors they induce. 

\begin{figure*}
    \centering
    \includegraphics[width=1.0\linewidth]{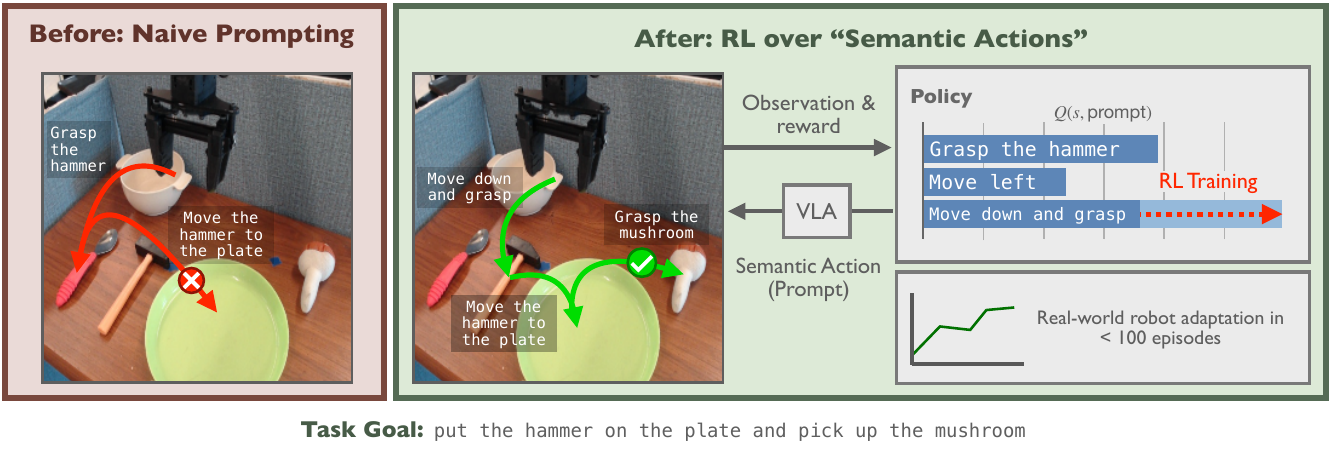}
    \caption{
    Steering VLA priors at deployment to solve complex, long-horizon tasks is challenging. To overcome this, we propose optimizing VLA language inputs via RL. The resulting \emph{semantic} action space probes skills already encoded in VLAs from pretraining, yielding expressive exploration and efficient adaptation. However, not all induced behaviors make task progress: by learning to decompose complex tasks into \emph{grounded} skills, our method identifies effective task-solving prompts and enables real-robot adaptation in under 100 episodes.
    }
    \label{fig:teaser}
    \vspace{-1.0em}
\end{figure*}

\section{Related Work}

\textbf{RL improvement of VLAs.}
There has been significant progress in developing general-purpose language-conditioned robot policies, such as vision-language-action models (VLAs) \cite{brohan2022rt,gu2023rt,team2024octo,kim2024openvla,black2024pi0,bjorck2025gr00t,team2025gemini,liu2025rdt,intelligence2025pi_,zha2026lap}.
While enabling effective task-solving in settings similar to their training data, VLAs often struggle to adapt to new tasks.
To address this, a variety of approaches for RL improvement of VLAs have been proposed. These include training a residual policy \citep{johannink2018residual,ankile2025imitation, yuan2024policy, julg2025refined, dong2025expo, xiao2025self, xu2026rl, sun2026prior}, steering the VLA's denoising process (when the VLA is a flow or diffusion model) \cite{wagenmaker2025steering}, augmenting the VLA with ``advantage conditioning'' \cite{amin2025pi}, and others \citep{mark2024policy, nakamoto2024steering, chen2025conrft, hu2025flare, guo2025improving, lu2025vla, liu2025can}. However, these works all run RL at the level of \emph{robot actions}, typically fixing the VLA's task prompt. In contrast, we run RL over the \emph{language prompt itself}, which we show unlocks novel task-solving abilities, enabling VLAs to generalize beyond single-task proficiency to solve novel long-horizon tasks. Our work is also related to hierarchical RL and skill learning \cite{daniel2016hierarchical,kulkarni2016hierarchical,peng2017deeploco,riedmiller2018learning,nachum2018data,gehring2021hierarchical,merel2018neural,ajay2020opal,singh2020parrot,pertsch2021accelerating,nasiriany2022learning,wilcoxson2024leveraging}, where a high-level policy learns to direct a low-level policy to maximize reward. While conceptually similar, unlike these works we focus on the setting where the ``low-level policy'' is a VLA and the interface between both policies is language.\loose

\textbf{Steering robots with language.}
Many works consider how a language-conditioned policy's prompt should be chosen for successful robot task-solving. Early works in this vein explore using LLMs or VLMs to generate useful instructions for solving real-world tasks
\cite{huang2022language, zeng2022socratic, huang2022inner, driess2023palm}, or rely on pre-programmed language-conditioned action primitives or ``skill'' policies as the language-to-robot interface \cite{huang2023voxposer, liang2023code, li2023interactive, zhang2023bootstrap, ha2023scaling}. More recently, several works have considered combining a VLM or LLM with a VLA or other language-conditioned policy, using the former to select the language prompts for the latter \cite{ahn2022can,shi2024yell,shi2025hi,chen2026steerable}.  Perhaps most similar to our work is  \cite{shah2025learning}---which selects a VLA's language command with a VLM, and utilizes the in-context learning ability of the VLM to adapt and improve the prompt based on past observations---and the work of \cite{kwok2026scaling}, which trains a prompt-action verifier using offline data, and then  selects the prompt to use online from a fixed set of initial prompts based on verifier score. 
However, none of these works incorporate RL or other forms of online improvement, which we find to be critical to achieving robust task-solving behavior.

\textbf{Language-based RL outside of robotics.}
While our focus is on robotics, a variety of other domains also seek to integrate RL and language. 
Our approach can be seen as a form of \emph{prompt optimization}, commonly used to elicit desired capabilities from language models \cite{shin2020autoprompt,zhou2022large,pryzant2023automatic}, and the most relevant of which are RL-based methods \cite{deng2022rlprompt,zhang2022tempera,jung2024discrete,wang2024promptagent,guo2023evoprompt,lu2022dynamic,jafari2024morl,batorski2025prl, kong2024prewrite, kwon2024stableprompt}. Outside of language models, language has also been used as a key tool in deep RL, enabling effective exploration \cite{mirchandani2021ella,mu2022improving,tam2022semantic,du2023guiding}, reward shaping \cite{carta2022eager,goyal2019using}, state abstraction design \cite{peng2024learning}, and more \cite{branavan2012learning,hermann2017grounded,misra2017mapping,narasimhan2018grounding}. Though these works seek to optimize language prompts with RL and use language as an abstraction to enable efficient learning, the specific challenges we face---in particular, the need for high sample-efficiency in real-world robotic deployments---necessitate a very different algorithmic strategy than those proposed here. \loose

\newcommand{\cO}{\mathcal{O}}
\newcommand{\cL}{\mathcal{L}}
\newcommand{\cM}{\mathcal{M}}
\newcommand{\cS}{\mathcal{S}}
\newcommand{\cA}{\mathcal{A}}
\newcommand{\cArob}{\mathcal{A}_{\mathrm{robot}}}
\newcommand{\cAlang}{\mathcal{A}_{\mathrm{sem}}}
\newcommand{\lang}{\ell}
\newcommand{\pivla}{\pi_{\mathrm{vla}}}
\newcommand{\pisemt}{\pi^t_{\mathrm{sem}}}
\newcommand{\pilang}{\pi_{\lang}}
\newcommand{\obs}{o}
\newcommand{\act}{a}
\newcommand{\avla}{a_{\mathrm{vla}}}
\newcommand{\bell}{\bm{\ell}}
\newcommand{\task}{\tau}
\newcommand{\tvlm}{t_{\mathrm{vlm}}}
\newcommand{\frakB}{\mathfrak{B}}
\newcommand{\Qbar}{\bar{Q}}
\newcommand{\traj}{\uptau}
\newcommand{\frakD}{\mathfrak{D}}
\newcommand{\Psemantic}{P_{\mathrm{sem}}}
\newcommand{\cMsemantic}{\cM_{\mathrm{sem}}}
\newcommand{\Qsem}{Q_{\mathrm{sem}}}
\newcommand{\Qbarsem}{\bar{Q}_{\mathrm{sem}}}

\section{Preliminaries}\label{sec:prelim}
\vspace{-0.5em}
\textbf{Markov decision processes.} We consider decision-making in Markov decision processes (MDPs), defined by a tuple $\cM = (\cS, \cA, P, P_0, \gamma, r)$, where $\cS$ is the set of states, $\cA$ the set of actions, $P : \cS \times \cA \rightarrow \triangle_{\cS}$ the transition function, $P_0 \in \triangle_{\cS}$ the initial state distribution, $\gamma \in [0,1]$ the discount factor, and $r : \cS  \rightarrow \mathbb{R}$ the reward function. An episode consists of a sequence of interactions between an agent and the environment. Each episode starts at initial state $s_0 \sim P_0$, the agent selects action $a_0$, receives reward $r(s_0)$, and transitions to state $s_1 \sim P(s_0, a_0)$, repeating this process until a terminal state is reached. A policy $\pi : \cS \rightarrow \triangle_{\cA}$ is a mapping from states to actions. 
The goal of RL is to learn a policy that maximizes the expected discounted reward, $V(\pi) := \mathbb{E}^\pi[\sum_{t} \gamma^t r(s_t)]$, where the expectation is over trajectories from rolling out $\pi$ in $\cM$. 
We define the $Q$-function as $Q^\pi(s, a) := \mathbb{E}^\pi [ \sum_{t} \gamma^t r(s_t) \,\Big|\, s_0 = s, a_0 = a]$, the expected discounted reward of taking action $a$ from state $s$ and then following policy $\pi$.

While usually taken as given, the parameterization of the action space $\cA$ is often itself a key design decision in RL. In robotic control, $\cA$ is most commonly the set of continuous robot actions (e.g., joint positions, motor torques, or end-effector deltas), which we generically denote as $\cArob$. However, as we will discuss in the following, correctly choosing $\cA$ can often enable much more efficient RL.

\textbf{Vision-language-action models (VLAs).}
Diverse large-scale robotic datasets $\frakD = \{ (\traj_t, \ell_t) \}_t$ often contain human-teleoperated trajectories, $\traj_t = (s_0^t, a_0^t, s_1^t, a_1^t, \ldots, )$ for $s_h^t \in \cS$ and $a_h^t \in \cArob$, paired with corresponding language commands, $\ell_t$. Behavioral cloning then trains a language-conditioned policy to produce action $a^t_h$ given state $s^t_h$ and language $\ell_t$ by ``copying'' the expert behaviors in the dataset $\frakD$. In principle, the resulting policies, given some task of interest described by a language command $\ell$, are able to produce actions $a_h \sim \pivla(\cdot \mid s_h, \ell)$ that lead the robot to complete the task. Such ``generalist'' policies, including ``vision-language-action'' models (VLAs), have shown significant promise in capturing language-controllable priors on effective robot behaviors, enabling general-purpose robotic task-solving abilities~\citep{lbm, black2024pi0, kim2024openvla, intelligence2025pi_}. 

\section{Leveraging VLAs as Semantic Action Priors for Efficient RL}
In this section, we outline our approach, \method, which leverages VLAs as controllable action priors that can be steered to enable effective online learning of complex tasks. In particular, we assume access to a VLA $\pivla$ and are interested in solving some task $\tau$ corresponding to a reward function $r$. As illustrated in Figure~\ref{fig:teaser}, \method treats the language prompt of a VLA as a ``semantic action'', and, through real-world interaction, learns to adaptively control this semantic action to steer the VLA to reliably complete tasks of interest.  

\subsection{VLAs as Semantically Controllable Action Priors}\label{sec:vla_prior}

\begin{wrapfigure}{r}{0.4\textwidth}
    \vspace{-1.5em}
    \centering
    \includegraphics[width=1.0\linewidth]{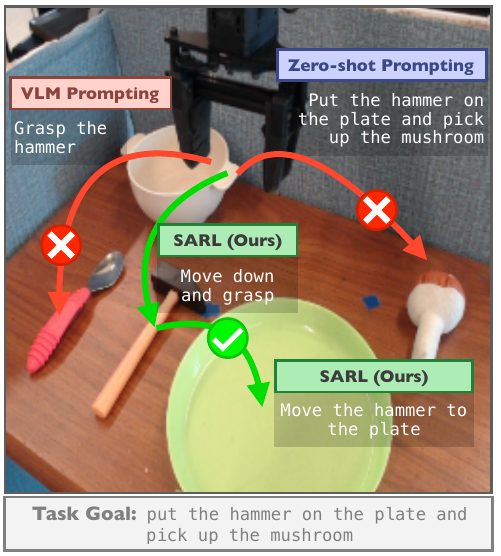}
    \caption{Adapting VLAs to solve new tasks requires \emph{decomposing} task-goals into achievable stages, and \emph{grounding} each stage in behaviors that can be executed successfully in the environment. VLAs prompted zero-shot struggle with both decomposition and grounding, and prompting VLAs with VLMs achieves decomposition but not grounding. Only \method achieves both by learning to optimize a VLA's prompt inputs with RL.} 
    \label{fig:grounding-and-decomposition}
    \vspace{-1.0em}
\end{wrapfigure}

In a typical VLA deployment, the user provides a fixed language command $\lang$ describing the task of interest $\task$, and then uses a VLA to execute this command. For each observation $s$, the VLA generates action $\act \sim \pivla(\cdot \mid s, \lang)$, which is executed on the robot. While effective in settings where the observation and language command are supported by the VLA's training data $\frakD$, this approach often fails to enable VLAs to solve new tasks not covered in the task repertoire spanned by $\frakD$.\loose

In this work, we take an alternative view on VLA deployment. While a VLA may fail to execute a complex or out-of-distribution (OOD) command zero-shot, it can often produce reliable behaviors if prompted with simple, physically grounded instructions \cite{chen2026steerable}. For instance, as illustrated in Figure~\ref{fig:grounding-and-decomposition}, the VLA fails to execute ``put the hammer on the plate and pick up the mushroom" because the command is too complex and because the hammer (likely an OOD object) is often confused for other objects. By contrast, ``move down and grasp" is simple and avoids referring to the hammer by name, successfully steering the robot. While prompting the VLA with the former instruction will fail to result in successful task completion, adaptively controlling the VLA by selecting a sequence of low-level, grounded instructions \emph{can} steer the robot to solve the overall task $\tau$. In other words, rather than viewing VLAs simply as policies to be statically prompted, we view them as \emph{semantically controllable action priors} that can be dynamically guided throughout deployment to accomplish complex tasks of interest.\loose

\textbf{Semantic MDPs.} 
This motivates a natural transformation of our original decision-making setting: rather than learning over the robot action space $\cArob$ as is typical in RL for robotics, we can learn in a \emph{semantic} action space---the space of language commands, which we denote as $\cAlang$---and deploy the VLA as a transformation between the two. Formally, rather than directly specifying action $a \in \cArob$, we can specify a ``semantic action'' $\ell \in \cAlang$, generate robot action $\avla \sim \pivla(\cdot \mid s, \ell)$ with our VLA, and transition to $s' \sim P(\cdot \mid s, \avla)$. We denote this induced transition function as $\Psemantic : \cS \times \cAlang \rightarrow \triangle_{\cS}$, and the corresponding induced ``semantic MDP'' as $\cMsemantic := (\cS, \cAlang, \Psemantic, P_0, \gamma, r)$.

By considering decision-making in $\cMsemantic$ instead of $\cM$, we shift the burden from directly specifying precise robot actions to  guiding the robot with semantic language commands, leveraging the VLA's priors to translate between semantic commands and robot actions. As described above, by correctly modulating these semantic actions, we can potentially steer the VLA to solve tasks that initially may appear outside its capabilities. Critically, learning over this semantic space enables efficient \emph{semantic exploration}---instead of exploring the entirety of $\cArob$, we can focus our exploration only on the ``semantically meaningful'' behaviors encoded in the VLA's priors and accessed via language prompting. As we will see, this enables highly efficient exploration and online improvement.

\subsection{\method: Learning in Semantic Action Spaces}
\label{subsec:SARL-algo}
Our proposed approach, \method, aims to learn how to correctly select these ``semantic actions'' in order to steer VLAs to task success. While a VLA may act as an effective prior over robot actions, in general we do not know which language commands will effectively steer it to solve tasks of interest, and we must learn this through interaction with both the VLA and the environment. While in principle this can be achieved by applying any RL algorithm to learn in the semantic MDP $\cMsemantic$, we propose a particular instantiation that seeks to enable efficient real-world learning.\loose

\textbf{\methodspelledout (\method).}
At a high level, \method operates by selecting a semantic action $\ell_t \in \cAlang$ at each step $t$ and state $s_t$, taking semantic action $\ell_t$ in $\cMsemantic$ (or, equivalently, generating action $a_t \sim \pivla(\cdot \mid s_t, \ell_t)$ and taking action $a_t$ in $\cM$), and observing reward $r_t$ and next state $s_{t+1}$. In order to determine which semantic action to take at each step, \method learns a \emph{semantic $Q$-function}, $\Qsem$, trained to estimate the effectiveness of semantic actions at making progress towards completing $\tau$ from each state $s$ through
temporal-difference (TD) backups~\cite{barto2021reinforcement}. When acting from state $s_t$, \method samples a semantic action from the softmax distribution induced by their $Q$-values, $\ell_t \sim \pisemt(\cdot \mid s)$ for $\pisemt(\ell \mid s) \propto \exp(\Qsem^t(s_t,\ell))$. A formal description of our approach can be found in Algorithm~\ref{alg:main}.

\newcommand{\Vbarsem}{\bar{V}_{\mathrm{sem}}}

\begin{figure}[H]
\vspace{-0.5em}
    \centering
    \begin{minipage}{\linewidth}
    \begin{algorithm}[H]
      \caption{\methodspelledout (\method)}
      \begin{algorithmic}[1]
        \State \textbf{input}: semantic environment $\cMsemantic$, semantic action space $\cAlang$
        \State Initialize $\Qsem^1$ randomly, replay buffer $\frakB \leftarrow \emptyset$
        \For{$t = 1, 2, 3, \ldots $}
        \State Set $\pisemt(\ell \mid s) \propto \exp(\Qsem^t(s,\ell))$
        \State Sample $\ell_t \sim \pisemt(\cdot \mid s_t)$
        \Statex \hspace{1em} {\color{blue} \texttt{// Equivalently, execute action $a_t \sim \pivla(\cdot \mid s_t, \ell_t)$ in $\cM$}}
        \State Execute semantic action $\ell_t$ in  $\cMsemantic$, observe reward $r_t$ and next state $s_{t+1}$ 
        \State $\frakB \leftarrow \frakB \cup \{ (s_t, \ell_t, r_t, s_{t+1}) \}$
        \State Set $\Vbarsem^t(s') \leftarrow \mathbb{E}_{\ell' \sim \pisemt(\cdot \mid s')} \big[ \Qbar_{\mathrm{sem}}^t(s', \ell') \big]$ and
        \begin{align*}
            \textstyle
            \Qsem^{t+1} \leftarrow \min_Q \sum_{(s,\ell,r,s') \in \frakB} (Q(s,\ell) - r - \Vbarsem^t(s'))^2 
        \end{align*}
        \EndFor
      \end{algorithmic}
    \label{alg:main}
\end{algorithm}
\vspace{-2em}
\end{minipage}
\end{figure}

\textbf{Compressing $\cAlang$ with VLMs.}
Na\"{\i}vely learning over the space of all possible VLA prompts ($\cAlang$) is intractable. To mitigate this, \method refines $\cAlang$ and reduces the search space by relying on large pretrained models, which can encode powerful semantic priors. In particular, vision-language models (VLMs) have proven effective at proposing semantic commands relevant for robotic control \cite{ahn2022can, chen2026steerable, shah2025learning}. Instead of searching over all possible prompts, \method provides a VLM with the current image observation $s_t$ and high-level task $\tau$, and prompts it to generate a set of \textit{candidate} semantic actions, $\cAlang^t$. $\cAlang^t$ is then used instead of $\cAlang$ in Algorithm~\ref{alg:main}, enabling \method to consider only a small set of candidate actions at each step. 

Note that, while VLMs can often generate reasonable candidate VLA prompts, they fundamentally lack grounding for which commands are actually effective. A potential command might appear reasonable to a VLM, yet the VLA's exact response to this command is highly dependent on the deployment setting, the VLA's training data, and other factors the VLM does not have access to. In other words, VLMs alone are not effective at directly controlling VLAs (as shown in \Cref{sec:experiments}). By interleaving VLM queries with real-world interaction, \method is able to achieve the best-of-both worlds---effectively leveraging the VLM's semantic priors, while also enabling improvement over these priors by grounding semantic actions in physical VLA behaviors from experience.

\section{Experiments}\label{sec:experiments}

\begin{figure*}
    \centering
    \vspace{-1.5em}
    \includegraphics[width=1\linewidth]{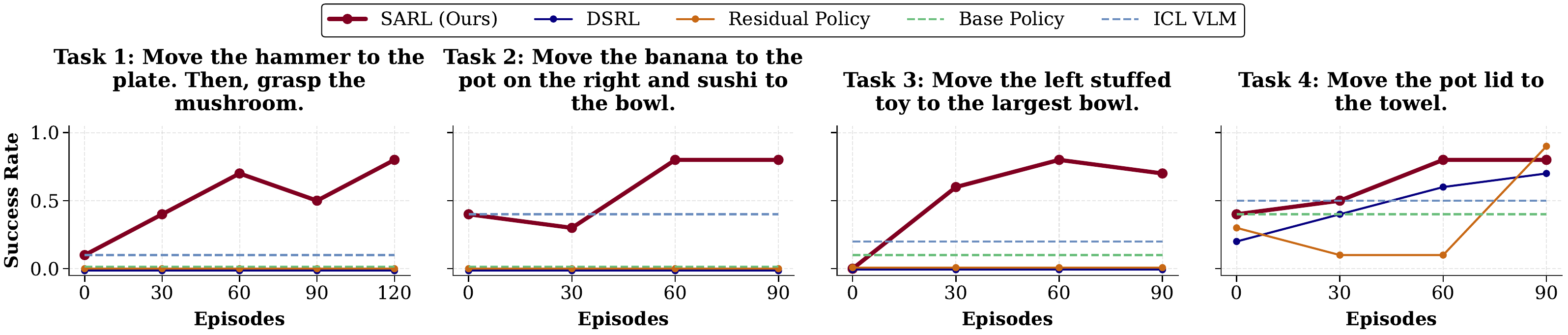}
    \caption{Across four complex, real-world tasks, learning over semantic actions with \method enables the best improvement of generalist policy behavior in deployment. In particular, \method enables new capabilities over prior steering methods such as DSRL \cite{wagenmaker2025steering} and Residual RL \cite{xiao2025self, ankile2025imitation, xu2026rl}, as these methods are fundamentally constrained by the base policy's performance under the task's original prompt. 
    In addition, \method outperforms an in-context learning VLM baseline (ICL VLM) \cite{chen2026steerable}, since VLMs struggle to ground semantic actions in physical behaviors. Each data-point represents $10$ evaluations.} 
    \label{fig:real-graphs}
\end{figure*}

\begin{figure*}
    \centering
    \includegraphics[width=1.0\linewidth]{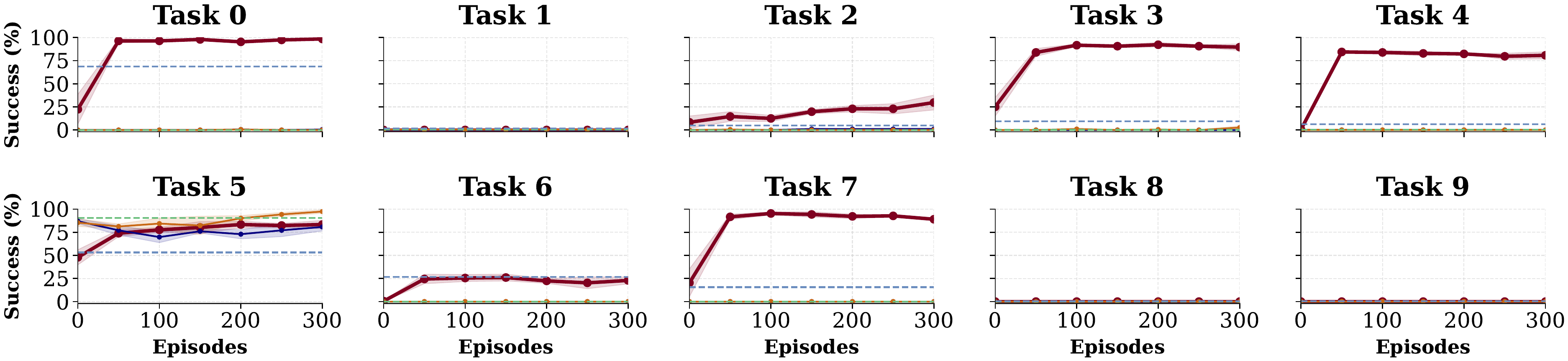}
    \caption{\method outperforms prior methods for deployment-time adaptation --- DSRL \cite{wagenmaker2025steering}, Residual RL \cite{xiao2025self, ankile2025imitation, xu2026rl}, and in-context learning with VLMs \cite{chen2026steerable} --- on long-horizon Libero-10 tasks \cite{liu2023libero}. \method successfully adapts a policy \cite{jain2025polaris} on five tasks, and matches performance on another already close to solved. Four tasks remain unsolved by any method. 
    Each graphed data-point represents $64$ evaluations and standard error over $3$ seeds.\loose}
    \label{fig:sim-graphs}
\end{figure*}

Our experiments evaluate \method both as a sample-efficient algorithm for adapting VLA priors and as a method for steering VLAs through high-level language instructions. We aim to answer three questions. First, can \method's semantic exploration space efficiently adapt VLAs to solve new tasks? Second, compared to traditional action-space RL methods,
does operating in a lifted, semantic prompt space allow \method to efficiently solve complex, long-horizon tasks where these prior methods may struggle? Finally, does optimizing prompts via online RL outperform non-RL VLA prompting methods (e.g., in-context learning, ICL, with VLMs), by explicitly grounding semantic instructions in the physical behaviors they induce? We investigate these questions across a suite of challenging multi-step tasks within both the Libero simulation benchmark \cite{liu2023libero} and with a real-world WidowX robot.\loose

\subsection{Experimental Setup}
\label{subsec:experimental-setup}

\begin{figure}
    \centering
    \includegraphics[width=1.0\linewidth]{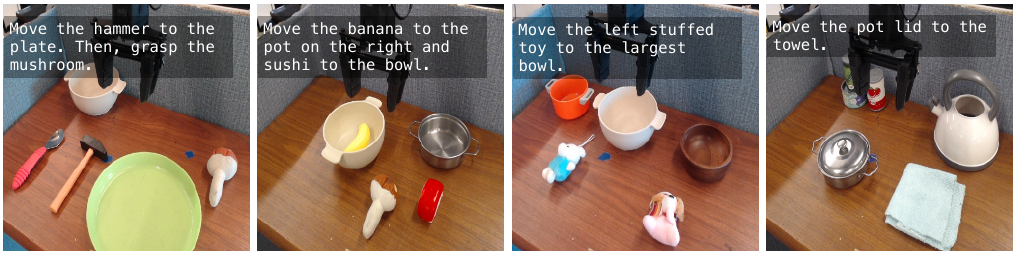}
    \caption{We test our approach on four complex, long-horizon tasks on the WidowX robot, illustrated above, along with ten tasks on the simulated Libero-10 benchmark \cite{liu2023libero}. 
    } 
    \label{fig:task-suite}
    \vspace{-1.5em}
\end{figure}

We evaluate \method on the simulated Libero-10 benchmark \cite{liu2023libero} and four challenging real-world WidowX tasks \cite{walke2023bridgedata} shown in Figure~\ref{fig:task-suite}. These tasks are specifically designed to be long-horizon, and involve multiple steps. In fact, the base VLA policies we use---based on $\pi_{0.5}$ \cite{intelligence2025pi_}, and finetuned on the Libero-90 \cite{liu2023libero} and Bridge v2 datasets \cite{walke2023bridgedata}---perform poorly on these tasks, only achieving significant performance on one task in each setting. Consequently, these tasks are a good evaluation domain for RL methods that leverage pre-trained generalist policies: while they contain many atomic behaviors that are well-represented in Libero and Bridge's training sets, they reveal the difficulty of applying these behaviors directly to complex new tasks. 

In addition to the original language prompt space for the VLA, in Libero we also introduce a ``reset-to-home'' command that resets the robot arm to its starting state. Note that this is straightforward to implement on a real robot (e.g. we simply reset the robot's joint positions), and the capacity to seamlessly integrate such auxiliary controllers highlights a unique strength of our approach.

We compare \method against two classes of baseline approaches. As VLA RL baselines, we consider \emph{Diffusion Steering via Reinforcement Learning (DSRL)} \cite{wagenmaker2025steering}, which learns to steer diffusion policies in their latent-noise space, and \emph{Residual RL} \cite{xiao2025self, ankile2025imitation, xu2026rl}, which learns a residual policy that predicts small corrective actions which are added to the base policy before execution. For both baselines we rely on standard implementations. To the best of our knowledge, all existing evaluations of DSRL and Residual RL use a \emph{fixed} VLA prompt. Therefore, for these comparisons, we provide these prior methods with the task prompts indicated in Figure~\ref{fig:task-suite} for real-world experiments, and the standard Libero task prompts in simulation.
For our second class of baselines, we explore approaches that adaptively adjust the VLA's prompt. In particular, we consider an in-context learning-based VLM approach that processes past observations in-context and uses this information to select prompts that steer the VLA to task completion \cite{chen2026steerable, shah2025learning}. For \method, we utilize the VLM-based restriction of $\cAlang$ outlined in Section~\ref{subsec:SARL-algo}. To simplify the prompt space further, the algorithm caches and reuses the first $N \in \{ 32, 64, 100 \}$ prompts proposed by the VLM. We find this to produce sufficiently expressive behaviors to enable effective learning. 

Finally, for real-world WidowX experiments, we collect three full demonstrations per task and seed them into \method and Residual RL's replay buffers prior to training. Because DSRL inherently cannot ingest such demonstrations---it learns over latent-noise space, rather than robot action space, where the demonstrations are given---it is evaluated without them. To collect these demonstrations, an operator merely provides the VLA with periodic language instructions. Note that the ease of collecting such demonstrations is a core advantage of language-based steering. Further experimental details are provided in Appendix~\ref{app:hyperparams}.

\subsection{\method Enables Efficient Adaptation of Generalist Policies}

We first evaluate \method's ability to learn effective task-solving behaviors compared to existing methods for improving VLAs in deployment. As shown in Figures~\ref{fig:real-graphs} and~\ref{fig:sim-graphs}, \method is able to significantly improve the performance of the base VLA across Libero-10 and the four long-horizon real-world tasks we consider---in both settings, \method is able to improve the VLA's initial success rate of near 0\% under the task prompt up to 80\% after only 60-100 online episodes. This suggests that the VLA's prompts are an expressive and high-quality space for exploration and task-learning, and that by modulating the language prompt, we can elicit task-solving behaviors that naive prompting fails to exhibit.  

Notably, our approach improves over traditional action-space RL methods, such as DSRL and Residual RL, which are unable to learn at all on several tasks. \method also significantly outperforms the ICL VLM baseline, which adaptively selects VLA prompts by in-context learning over a history of interactions. These results highlight \method's effectiveness as a method for adaptation on real-world robots, enabling novel task-solving abilities in VLAs. 

We emphasize that the tasks in our evaluation suite are unique in their complexity and horizon for deployment-time adaptation compared to prior works \cite{wagenmaker2025steering, xu2026rl, xiao2025self}. Our tasks, which by design have multiple stages and require invoking sequences of skills seen during training on Bridge v2 (real experiments) \cite{walke2023bridgedata} and Libero (sim experiments) \cite{liu2023libero}, are not able to be solved by action-space steering methods, such as DSRL and Residual RL. The unique ability of \method to solve these tasks highlights that it unlocks a completely new capability over these prior steering methods.
\vspace{-0.5em}

\subsection{\method vs Action-Space RL: Editing Prompts Unlocks Fundamentally New Adaptation Abilities}

\begin{figure*}
    \vspace{-1.5em}
    \centering
    \includegraphics[width=1.0\linewidth]{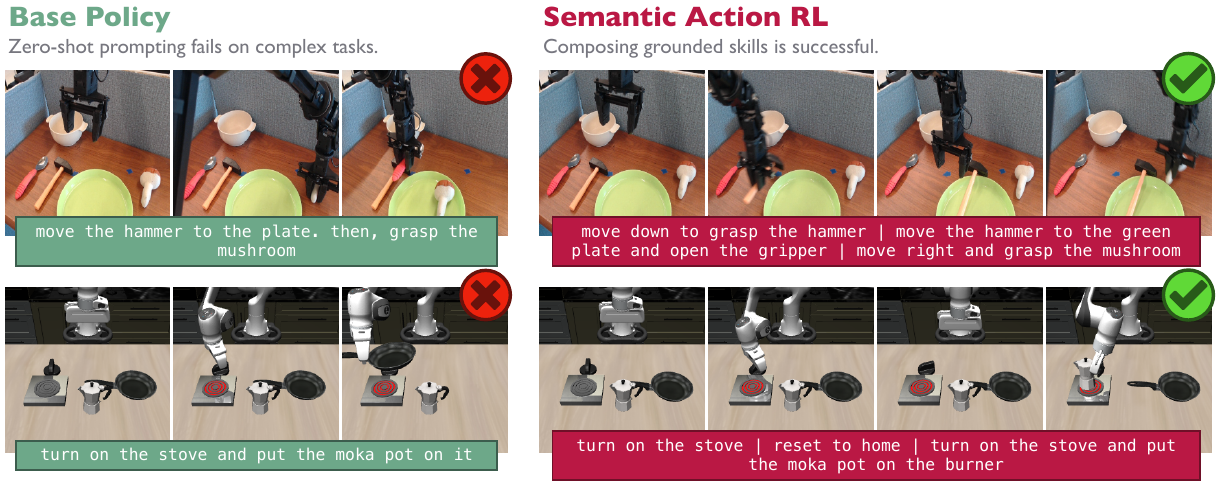}
    \caption{On complex, long-horizon tasks, the base policy fails when zero-shot prompted. \method solves them through learned prompting, sequencing skills covered under the pretraining distribution.}
    \label{fig:base-vs-sarl}
    \vspace{-1.0em}
\end{figure*}

As shown in Figures~\ref{fig:real-graphs} and~\ref{fig:sim-graphs}, \method significantly outperforms standard RL steering methods (DSRL and Residual RL) that operate over robot actions directly. We hypothesize that this is because (1) existing steering methods are constrained to a single task prompt, and (2) the VLA's action distribution under this task prompt does not adequately cover the diversity of actions needed to solve the task, making action-based steering ineffective. 

Indeed, when evaluated on our multi-step task suite and prompted zero-shot, the base policy's action distribution tends to collapse into entirely incorrect modes, far from an optimal policy's distribution. A qualitative example is shown in Figure~\ref{fig:base-vs-sarl}, where a VLA instructed to ``move the hammer to the plate. then, grasp the mushroom'' (Row 1), instead moves the mushroom to the plate and grabs a spoon. 

This is catastrophic for methods that steer over actions directly. For instance, DSRL can filter to find good actions from the base policy's distribution, but cannot synthesize fundamentally new ones. Similarly, residual RL is restricted to exploring a narrow funnel around the base policy's actions. Consequently, these baselines only learn effectively when the base policy already exhibits close to successful behaviors on the underlying task (e.g., Libero Task 5 or real-world Task 4), but cannot recover from the significantly incorrect behavior modes observed when task prompts are out-of-support.
In contrast, \method is not constrained in this way. By modulating prompts, \method explores regions of the VLA’s behavioral prior that remain entirely inaccessible to action-only steering methods. This capability is critical for solving long-horizon tasks, and \method uniquely drives significant progress across five Libero-10 tasks, along with all tested WidowX tasks. 

\subsection{\method vs VLM Prompting: Instruction Grounding is as Important as Decomposition}

\begin{figure*}
    \vspace{-1.5em}
    \centering
    \includegraphics[width=1.0\linewidth]{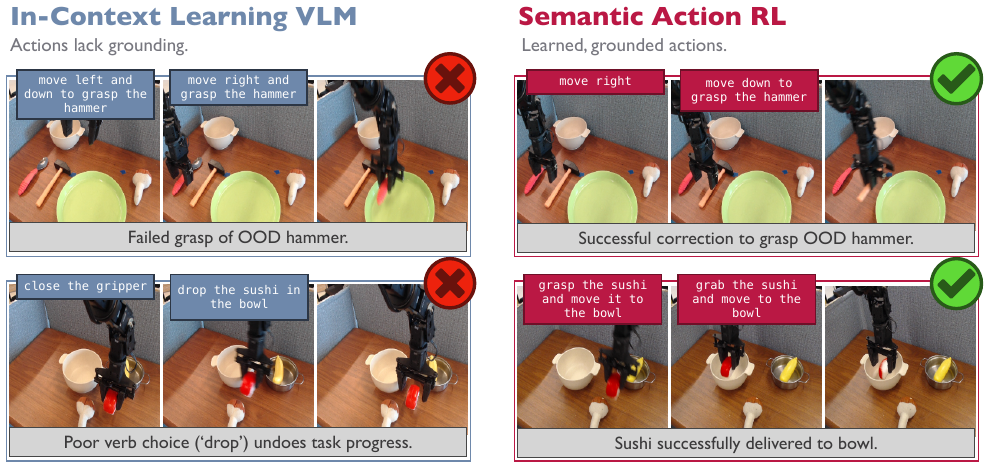}
    \caption{Across tasks, an in-context learning VLM \cite{chen2026steerable} selects instructions that are semantically meaningful, but lack grounding in physical behaviors, leading to task failures. By contrast, \method's learned semantic actions correctly navigate scenarios where choosing the right instruction is critical. Additional examples can be found in Figure~\ref{fig:vlm-vs-sarl-full}.}
    \label{fig:vlm-vs-sarl}
    \vspace{-1.5em}
\end{figure*}

Our results show that \method enables highly efficient learning of complex, long-horizon tasks even when task-prompts fall outside the base policy's support. To disentangle whether this capability stems from more than just instruction decomposition (ie. breaking down task prompts into sequences of short-horizon commands), we consider a closer comparison of our approach against its VLM baseline \cite{chen2026steerable}, which selects instructions via in-context learning (ICL) over interaction history. 

As shown in Figures~\ref{fig:real-graphs} and~\ref{fig:sim-graphs}, the VLM baseline outperforms the zero-shot base policy, confirming that task decomposition does improve performance. However, across the board, \method significantly outperforms the VLM. Qualitative examples in Figure~\ref{fig:vlm-vs-sarl} illustrate why: the VLM frequently selects commands that seem semantically plausible but still cause execution failures. For instance, in Row 1, the VLM is unable to pick the hammer (likely an OOD object): referencing the ``hammer" by name mistakenly causes the VLA to grasp a spoon, and only spatial commands (e.g., ``move right'') successfully guide the VLA. In another instance (Row 2), the VLM selects an incorrect verb (``drop''), causing the VLA to prematurely release a piece of sushi before reaching the target. While the VLM can learn in-context, it cannot recover from such failures in a single episode. By contrast, through many episodes of experience, \method is able to learn the grounded behavior induced by each language command, and uses this knowledge to select language commands appropriately. Our results highlight that this ability is critical for performant adaptation.

\section{Discussion and Limitations}

We proposed \method, a new method for steering VLAs over their language inputs. Our results illustrate that \method enables highly efficient learning, particularly excelling at complex, long-horizon tasks where standard steering methods are unsuccessful. We additionally disentangle that \method's performance comes from both semantically decomposing task goals using a VLM, while also learning to ground those behaviors in the actions they induce through experience. We find \method to be a powerful yet efficient approach for adapting generalist robot policies online, unlocking new capabilities over prior methods. 

The two main limitations of \method are speed (due to having a VLM in the loop), and the prerequisite of having VLAs that produce diverse behaviors when prompted with diverse instructions. For the former, an interesting extension of our work is studying whether VLMs are required at deployment time, or if the learned $Q$-function can generalize. For the latter, we believe the community will continue to develop stronger language-conditioned policies, making our work increasingly relevant. We hope that \method will inspire the training of policies with better language following capabilities, enable the creation of even stronger adaptation methods by combining semantic-action space steering with complementary robot-action space steering, and eventually, in a future world with powerful and general VLAs, be a part of a standard toolkit that enables robots to autonomously adapt and solve problems in open-ended environments.

\bibliography{main}  

\appendix
\clearpage
\section{Experimental Details}
\label{app:hyperparams}

\subsection{\method: In-practice Optimizations}
We make three optimizations to Algorithm~\ref{alg:main} presented in Section~\ref{subsec:SARL-algo} to improve learnability and speed.

First, as discussed in Section~\ref{subsec:SARL-algo}, na\"{\i}vely learning over the space of all possible VLA prompts ($\cAlang$) is intractable. Instead, at every execution step, we use a VLM to produce a set of candidate semantic actions $\cAlang^t$, which \method is then run over. However, since the VLM can still, in principle, generate any language prompt, this approach requires a prompt featurization scheme that can both 1) encode the semantic meaning of each prompt 2) generalize the VLA's grounded behavior resulting from each prompt across these features. In practice, we find the latter to be especially difficult with off-the-shelf embeddings. 

As an alternate approach, we cache the first $k$ prompts seen by the algorithm (either generated by the VLM, or provided by a human during demonstrations). After the cache has been filled, the VLM is queried to generate $\cAlang^t$, but may only choose commands from the size-$k$ cache, rather than generating them open-endedly. This allows for a one-hot prompt featurization which we find enables efficient learning. $|\cAlang^t|$ and $k$ for each task can be found in Table~\ref{tab:sarl_hyperparams} as \texttt{outer\_policy.max\_candidates} and \texttt{outer\_policy.cache\_size}, respectively.

We consider two additional optimizations: First, for each next state $s_{t+1}$ added to the replay buffer in Algorithm~\ref{alg:main}, we cache the corresponding generated $\cAlang^{t+1}$, which avoids the need to call the VLM during Temporal-Difference (TD) backups \cite{barto2021reinforcement}. Second, instead of generating $\cAlang^t$ for each $t$, we use the same $\cAlang^t$ for $\tvlm$ steps before regenerating --- this speeds up the time to collect each episode. A revised Algorithm~\ref{alg:main} with these optimizations is provided in Algorithm~\ref{alg:opt}.

\newcommand{\cC}{\mathcal{C}}
\newcommand{\cAlangnew}{\cAlang^{\mathrm{new}}}
\newcommand{\cAlangprev}{\cAlang^{\mathrm{prev}}}

\begin{figure*}[t!]
\vspace{-1.5em}
    \centering
    \begin{minipage}{\linewidth}
    \begin{algorithm}[H]
      \caption{\methodspelledout (\method) with Optimizations}
      \begin{algorithmic}[1]
        \State \textbf{input}: semantic environment $\cMsemantic$, semantic action set $\cAlang$, cache size limit $k$, regeneration period $\tvlm$, demonstrations $\frakD$ (optional).
        \State Initialize $\Qsem^1$ randomly, replay buffer $\frakB \leftarrow \emptyset$, prompt cache $\cC \leftarrow \emptyset$. If demonstrations $\frakD$ are provided, update $\frakB$ and $\cC$ accordingly. 
        
        \Statex
        \Function{GetCandidates}{$s, t, \cC, \cAlangprev$}
          \If{$\mathtt{mod}(t - 1, \tvlm) == 0$} 
            \Statex \hspace{2.6em} {\color{blue} \texttt{// Refresh candidate actions}}
            \If{$|\cC| < k$} 
              \Statex \hspace{4em} {\color{blue} \texttt{// Cache not yet full: generate open-endedly and cache}}
              \State Query VLM open-endedly for candidate prompts $\cAlangnew$ given $s$
              \State $\cC \leftarrow \cC \cup \cAlangnew$ 
            \Else 
              \Statex \hspace{4em} {\color{blue} \texttt{// Cache is full: restrict VLM choices to the cache}}
              \State Query VLM to select candidate prompts $\cAlangnew \subseteq \cC$ given $s$
            \EndIf
            \State \Return $\cAlangnew$
          \Else 
            \Statex \hspace{2.6em} {\color{blue} \texttt{// Reuse previous step's candidates}}
            \State \Return $\cAlangprev$
          \EndIf
        \EndFunction
        \Statex
        
        \State $\cAlang^1 \leftarrow \textsc{GetCandidates}(s_1, 1, \cC, \emptyset)$ 
        \For{$t = 1, 2, 3, \ldots $}
        \State Set $\pisemt(\ell \mid s_t ; \cAlang^t) \propto \exp(\Qsem^t(s_t,\ell))$ for $\ell \in \cAlang^t$
        \State Sample $\ell_t \sim \pisemt(\cdot \mid s_t ; \cAlang^t)$
        \Statex \hspace{1em} {\color{blue} \texttt{// Equivalently, execute action $a_t \sim \pivla(\cdot \mid s_t, \ell_t)$ in $\cM$}}
        \State Execute semantic action $\ell_t$ in $\cMsemantic$, observe reward $r_t$ and next state $s_{t+1}$ 
        \State $\cAlang^{t+1} \leftarrow \textsc{GetCandidates}(s_{t+1}, t+1, \cC, \cAlang^t)$ 
        \Statex \hspace{1em} {\color{blue} \texttt{// Cache $\cAlang^{t+1}$ to eliminate VLM calls during TD backups}}
        \State $\frakB \leftarrow \frakB \cup \{ (s_t, \ell_t, r_t, s_{t+1}, \cAlang^{t+1}) \}$ 

        \State Set $\Vbarsem^t(s') \leftarrow \mathbb{E}_{\ell' \sim \pisemt(\cdot \mid s' ; \cAlang')} \big[ \Qbar_{\mathrm{sem}}^t(s', \ell') \big]$ where  $\pisemt(\ell' \mid s'; \cAlang') \propto \exp(\Qsem^t(s', \ell'))$ for $\ell' \in \cAlang'$ and
        \begin{align*}
            \textstyle
            \Qsem^{t+1} \leftarrow \min_Q \sum_{(s,\ell,r,s',\cAlang') \in \frakB} (Q(s,\ell) - r - \Vbarsem^t(s'))^2 
        \end{align*}
        \EndFor
      \end{algorithmic}
    \label{alg:opt}
\end{algorithm}
\end{minipage}
\end{figure*}

\subsection{MDP Definition}

For steering methods that make decisions over chunks of low-level actions, it is helpful to distinguish two distinct Markov Decision Processes (MDPs): low-level MDPs and high-level MDPs. In low-level MDPs, each transition corresponds to executing a single action in the underlying environment. By contrast, high-level MDPs operate over a coarser temporal resolution, where each transition corresponds to a ``chunk'' of low-level actions executed entirely open-loop. When converting low-level MDPs into high-level MDPs, we define a \textit{high-level step size} to be the number of low-level actions executed open-loop in each transition of the high-level MDP. The horizon of a task in a high-level MDP is therefore its horizon in the corresponding low-level MDP divided by the high-level step-size. 

The methods evaluated in this work operate over high-level MDPs, which improves rollout speed (especially when querying a VLM is required at each MDP step) and temporal consistency of actions. As shown in Table~\ref{tab:mdp-def}, each task's horizon --- defined in terms of low-level environment steps --- remains constant across all compared methods. However, the high-level step size varies across tasks and methods.

For real-world WidowX experiments, we use a high-level step size of $20$ for the VLM baseline, querying the VLM every $20$ environment steps for a new prompt to issue the VLA. This keeps in line with \citet{chen2026steerable}, which this baseline is based on. For our method, \method, we query the VLM at the same rate, every $20$ steps ($\tvlm=20$), to generate candidate semantic actions $\cAlang^t$. However, we use a high-level step size of $1$, allowing the RL policy fine-grained control to choose which of these candidates is most relevant at each state. For a fair comparison, we also set the high-level step size of DSRL \cite{wagenmaker2025steering} and Residual RL baselines \cite{xiao2025self, ankile2025imitation, xu2026rl} to be $1$, so that they are also able to make reactive, fine-grained adjustments based on the state.

In simulation, we sweep the VLM's high-level step size over $40$, $80$ and $120$ and find $40$ to be best. We additionally sweep \method's high-level step size over $40$ and $120$ (keeping $\tvlm=1$) and find better mode-consistent exploration with $120$. The original DSRL implementation \cite{wagenmaker2025steering} uses $\pi_0$ \cite{black2024pi0} with a high-level step size of $20$. Our work uses $\pi_{0.5}$ \cite{intelligence2025pi_} --- for improved language-following capabilities --- with a high-level step size of $10$, as this is the action horizon of $\pi_{0.5}$. For fairness, we use the same step size for Residual RL. We find our hyperparameter selection for both DSRL and Residual RL to be competitive with the Libero \cite{liu2023libero} results from \citet{wagenmaker2025steering}. Further details on hyperparamters for DSRL \cite{wagenmaker2025steering}, Residual RL \cite{xiao2025self, ankile2025imitation, xu2026rl}, and \method can be found in Tables~\ref{tab:dsrl_hyperparams},~\ref{tab:residual_rl_hyperparams}, and~\ref{tab:sarl_hyperparams}, respectively.

\begin{table*}[h!]
\caption{\textbf{MDP Definition Across Tasks and Methods.} Task horizon is the total number of low-level environment steps allocated per task. The high-level step size defines the conversion between high-level steering steps to low-level environment steps. To get the horizon $H$ of the high-level MDP for each method and task (over which learning happens), divide the task horizon by the high-level step size.}
\label{tab:mdp-def}
\centering
\begin{tabular}{p{4.5cm} cccc cccc}
\toprule
& \multicolumn{4}{c}{\textbf{\texttt{Task Horizon}}} & \multicolumn{4}{c}{\textbf{\texttt{High-Level Step Size}}} \\
\cmidrule(lr){2-5} \cmidrule(lr){6-9}
\textbf{Task} & \small{\textbf{\method}} & \small{\textbf{DSRL}} & \small{\textbf{Residual}} & \small{\textbf{VLM}} & \small{\textbf{\method}} & \small{\textbf{DSRL}} & \small{\textbf{Residual}} & \small{\textbf{VLM}} \\
\midrule
\textbf{Real Task 1:} Move the hammer to the plate. Then, grasp the mushroom. & 120 & 120 & 120 & 120 & 1 & 1 & 1 & 20 \\
\addlinespace
\textbf{Real Task 2:} Move the banana to the pot on the right and sushi to the bowl. & 160 & 160 & 160 & 160 & 1 & 1 & 1 & 20 \\
\addlinespace
\textbf{Real Task 3:} Move the left stuffed toy to the largest bowl. & 100 & 100 & 100 & 100 & 1 & 1 & 1 & 20 \\
\addlinespace
\textbf{Real Task 4:} Move the pot lid to the towel. & 60 & 60 & 60 & 60 & 1 & 1 & 1 & 20 \\
\addlinespace
\textbf{Libero-10} & 600 & 600 & 600 & 600 & 120 & 10 & 10 & 40 \\
\bottomrule
\end{tabular}
\end{table*}

\subsection{Reset-to-home}

In simulation, we evaluate \method along with baselines on the Libero-10 (Libero-long) benchmark \cite{liu2023libero}. As mentioned in Section~\ref{subsec:experimental-setup}, we use a VLA based on $\pi_{0.5}$ \cite{intelligence2025pi_}, and finetuned on the Libero-90 and DROID datasets \cite{liu2023libero, khazatsky2024droid}. In addition to the original language prompt space for this VLA, we also introduce a ``reset-to-home'' command that resets the robot arm to its starting state. We find such a command to be necessary to chain individual skills and make progress on the long-horizon tasks of the Libero-10 benchmark. Note that the ``reset-to-home" functionality is straightforward to implement on both simulated and real robots, since it simply runs a PID controller to match the robot's joint-angles or end-effector pose at its starting configuration. Note that the capacity to seamlessly integrate such auxiliary controllers through language highlights a unique strength of our approach. Both \method and the VLM baseline may leverage the ``reset-to-home'' command, and it is up to the individual approach how many steps to run this instruction for.

\subsection{Environment Step Plots}

\begin{figure*}
    \centering
    \includegraphics[width=1\linewidth]{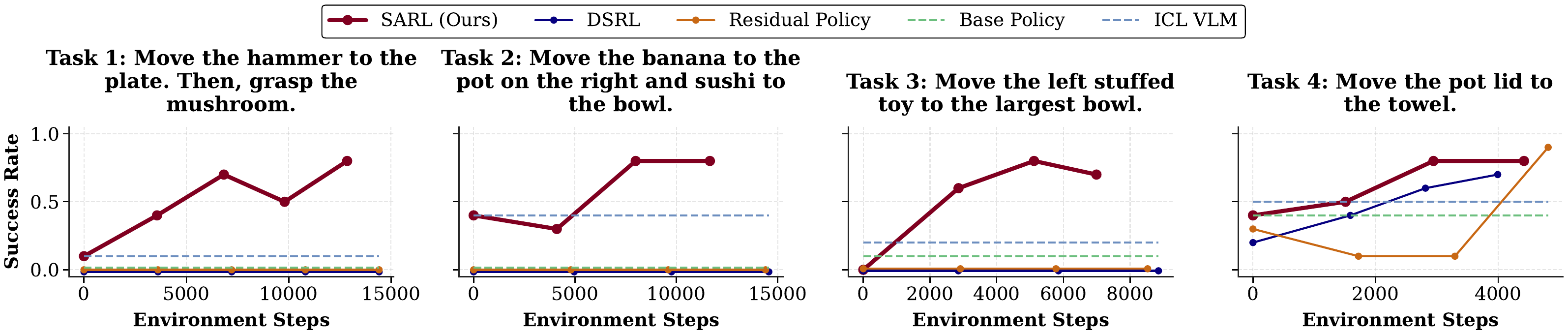}
    \caption{Across four complex, real-world tasks, learning over semantic actions with \method enables the best improvement of generalist policy behavior in deployment. \method enables new capabilities over prior steering methods such as DSRL \cite{wagenmaker2025steering} and Residual RL \cite{xiao2025self, ankile2025imitation, xu2026rl}, as these methods are fundamentally constrained by the base policy's performance under the task's original prompt. In particular, both DSRL and Residual RL perform best on Task 4, which has the least complex task goal. 
    Finally, our method outperforms an in-context learning VLM baseline (ICL VLM) \cite{chen2026steerable}, since VLMs struggle to ground semantic actions in physical behaviors. Each data-point represents $10$ evaluations. This plot mirrors Figure~\ref{fig:real-graphs}, but the x-axis is changed to measure environment steps instead of episodes.} 
    \label{fig:real-graphs-env-steps}
    \vspace{-0.5em}
\end{figure*}

In the main paper, we provide plots which measure performance against the number of episodes spent training. In Figure~\ref{fig:real-graphs-env-steps}, we provide the same plot with the x-axis swapped to show the number of environment steps spent training, as this is a standard choice in many prior works.

\subsection{VLM Prompts}

We use the Gemini model family \cite{team2023gemini, team2025gemini} for all VLM calls in \method and the VLM baseline. All VLM calls use the \texttt{gemini-3-flash-preview} model and are allocated $1024$ thinking tokens. Our methods use six total prompts (across real and sim experiments, two prompts generate candidate semantic actions $\cAlang^t$ open-endedly and from a cache, and another is used for the in-context learning VLM baseline). All prompts make minimal modifications to \citet{chen2026steerable}. Prompts for real are provided in Figures~\ref{fig:real-open-ended},~\ref{fig:real-cache}, and ~\ref{fig:real-sp-baseline} and those for sim are provided in Figures~\ref{fig:sim-open-ended},~\ref{fig:sim-cache}, and ~\ref{fig:sim-sp-baseline}.

\subsection{Additional Qualitative Examples}

\begin{figure*}[h!]
    \centering
    \includegraphics[width=1.0\linewidth]{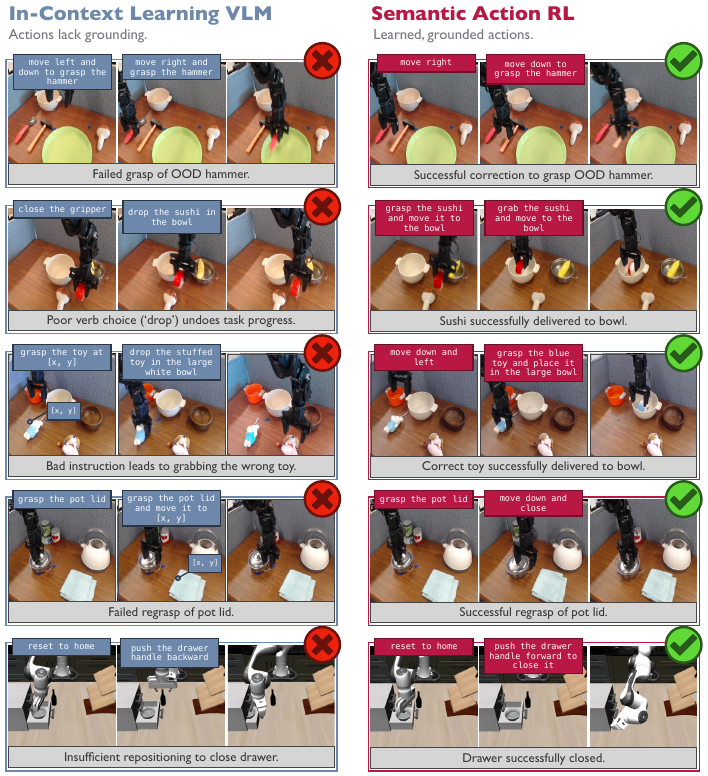}
    \caption{Across tasks, an in-context learning VLM \cite{chen2026steerable} takes semantic actions that are semantically meaningful, but lack grounding in physical behaviors, leading to task failures. By contrast, \method's learned semantic actions correctly navigate scenarios where choosing the right instruction is critical.}
    \label{fig:vlm-vs-sarl-full}
\end{figure*}

Additional qualitative examples of catastrophic failures due to the in-context learning VLM baseline's lack of grounding are shown in Figure~\ref{fig:vlm-vs-sarl-full} (extending Figure~\ref{fig:vlm-vs-sarl}).

\begin{table*}[htbp]
\caption{\textbf{DSRL Hyperparameters.} Implementations make minimal modifications to \citet{wagenmaker2025steering}. \textit{Real} indicates experiments run on the WidowX while \textit{Sim} indicates experiments run on Libero-10 \cite{liu2023libero}.}
\label{tab:dsrl_hyperparams}
\centering
\small
\begin{tabular}{llp{5.5cm}}
\toprule
\textbf{Hyperparameter} & \textbf{Value} & \textbf{Description} \\
\midrule
\texttt{outer\_policy.action\_dim} & 32 & Action space dimension, which is the same as the latent-noise dim for $\pi$ series policies \cite{intelligence2025pi_}. This low-dim action is repeated across the predicted action horizon as done in \cite{wagenmaker2025steering} \\
\texttt{outer\_policy.state\_dim} & 8 & Policy has access to a proprioceptive observation \\
\texttt{outer\_policy.policy\_chunk\_size} & \begin{tabular}[t]{@{}l|l@{}} \textit{Real:} 1 & \textit{Sim:} 10 \end{tabular} & Number of actions used from each evaluation of $\pi$ series policies \cite{intelligence2025pi_} which generate chunks of actions  \\
\texttt{outer\_policy.image\_resolution} & [64, 64] & Image resolution \\
\texttt{outer\_policy.camera\_names} & [``global"] & We only use scene cameras for all experiments \\
\texttt{env.image\_resolution} & [256, 256] & Image resolution of the base environment's camera \\
\texttt{sac.batch\_size} & 256 & Optimization batch size for the replay buffer \\
\texttt{sac.actor\_lr} & 0.0001 & Learning rate for the SAC \cite{haarnoja2018soft} actor network \\
\texttt{sac.critic\_lr} & 0.0003 & Learning rate for the SAC critic network \\
\texttt{sac.temp\_lr} & 0.0003 & Learning rate for the SAC entropy coefficient ($\alpha$) \\
\texttt{sac.hidden\_dims} & [128, 128, 128] & Hidden dimensions for SAC MLP layers \\
\texttt{sac.cnn\_features} & [32, 32, 32, 32] & Feature maps per layer for the CNN encoder \\
\texttt{sac.cnn\_strides} & [2, 2, 2, 2] & Strides per layer for the CNN encoder \\
\texttt{sac.cnn\_padding} & \texttt{VALID} & Padding type for the CNN encoder layers \\
\texttt{sac.cnn\_latent\_dim} & 50 & CNN output feature dim \\
\texttt{sac.tau} & 0.005 & Target network soft-update rate \\
\texttt{sac.critic\_reduction} & \texttt{min} & Ensemble reduction strategy for target Q-values \\
\texttt{sac.num\_qs} & 10 & Number of Q-networks in the critic ensemble \\
\texttt{sac.dropout\_rate} & 0.0 & Dropout rate applied inside SAC networks \\
\texttt{sac.init\_temperature} & 1.0 & Initial value for the SAC entropy coefficient \\
\texttt{sac.target\_entropy} & -16.0 & Target entropy \\
\texttt{sac.action\_range} & [-1.0, 1.0] & Defines the allowed min/max action. Note: SAC always sees actions in [-1.0, 1.0], but actions are scaled to be within \texttt{sac.action\_range} before executing in the environment. \\
\texttt{sac.discount} & $1 - \frac{1}{H}$ & Discount is defined based on the task horizon $H$ --- see Table~\ref{tab:mdp-def}. \\
\texttt{train.learning\_starts} & \begin{tabular}[t]{@{}l|l@{}} \textit{Real:} 10 & \textit{Sim:} 25 \end{tabular} & Episodes collected via uniformly sampling \texttt{sac. action\_range} to seed the replay buffer pre-learning  \\
\texttt{train.multi\_grad\_step} & \begin{tabular}[t]{@{}l|l@{}} \textit{Real:} 10 & \textit{Sim:} 20 \end{tabular} & Gradient steps per transition added to the buffer \\
\texttt{train.offline\_multi\_grad\_step} & \begin{tabular}[t]{@{}l|l@{}} \textit{Real:} 0 & \textit{Sim:} 5 \end{tabular} & Gradient steps per transition added to the buffer for episodes collected before \texttt{train. learning\_starts} \\
\texttt{eval.eval\_every} & \begin{tabular}[t]{@{}l|l@{}} \textit{Real:} 30 & \textit{Sim:} 50 \end{tabular} & Number of training episodes between evaluations \\
\texttt{eval.num\_evaluations} & \begin{tabular}[t]{@{}l|l@{}} \textit{Real:} 10 & \textit{Sim:} 64 \end{tabular} & Number of evaluation episodes \\
\bottomrule
\end{tabular}
\end{table*}

\begin{table*}[htbp]
\caption{\textbf{Residual RL Hyperparameters.} Implementations make minimal modifications to \citet{wagenmaker2025steering} and following standard Residual RL implementations \cite{xiao2025self, ankile2025imitation, xu2026rl}. \textit{Real} indicates experiments run on the WidowX while \textit{Sim} indicates experiments run on Libero-10 \cite{liu2023libero}.}
\label{tab:residual_rl_hyperparams}
\centering
\small
\begin{tabular}{llp{5.5cm}}
\toprule
\textbf{Hyperparameter} & \textbf{Value} & \textbf{Description} \\
\midrule
\texttt{outer\_policy.action\_dim} & \begin{tabular}[t]{@{}l|l@{}} \textit{Real:} 7 & \textit{Sim:} 70 \end{tabular} & Action space dimension, which is $7 \cdot \text{\texttt{outer\_policy.policy\_chunk\_size}}$, since Residual RL predicts a correction for each action to be executed \\
\texttt{outer\_policy.state\_dim} & \begin{tabular}[t]{@{}l|l@{}} \textit{Real:} 15 & \textit{Sim:} 78 \end{tabular} & Policy has access to an 8-dim proprioceptive observation along with knowledge of the base policy's action \\
\texttt{outer\_policy.policy\_chunk\_size} & \begin{tabular}[t]{@{}l|l@{}} \textit{Real:} 1 & \textit{Sim:} 10 \end{tabular} & Number of actions used from each evaluation of $\pi$ series policies \cite{intelligence2025pi_} which generate chunks of actions  \\
\texttt{outer\_policy.image\_resolution} & [64, 64] & Image resolution \\
\texttt{outer\_policy.camera\_names} & [``global"] & We only use scene cameras for all experiments \\
\texttt{env.image\_resolution} & [256, 256] & Image resolution of the base environment's camera \\
\texttt{sac.batch\_size} & 256 & Optimization batch size for the replay buffer \\
\texttt{sac.actor\_lr} & 0.0001 & Learning rate for the SAC \cite{haarnoja2018soft} actor network \\
\texttt{sac.critic\_lr} & 0.0003 & Learning rate for the SAC critic network \\
\texttt{sac.temp\_lr} & 0.0003 & Learning rate for the SAC entropy coefficient ($\alpha$) \\
\texttt{sac.hidden\_dims} & [128, 128, 128] & Hidden dimensions for SAC MLP layers \\
\texttt{sac.cnn\_features} & [32, 32, 32, 32] & Feature maps per layer for the CNN encoder \\
\texttt{sac.cnn\_strides} & [2, 2, 2, 2] & Strides per layer for the CNN encoder \\
\texttt{sac.cnn\_padding} & \texttt{VALID} & Padding type for the CNN encoder layers \\
\texttt{sac.cnn\_latent\_dim} & 50 & CNN output feature dim \\
\texttt{sac.tau} & 0.005 & Target network soft-update rate \\
\texttt{sac.critic\_reduction} & \texttt{min} & Ensemble reduction strategy for target Q-values \\
\texttt{sac.num\_qs} & 10 & Number of Q-networks in the critic ensemble \\
\texttt{sac.dropout\_rate} & 0.0 & Dropout rate applied inside SAC networks \\
\texttt{sac.init\_temperature} & 1.0 & Initial value for the SAC entropy coefficient \\
\texttt{sac.target\_entropy} & $-\text{\texttt{action\_dim}} / 2$ & Target entropy \\
\texttt{sac.action\_range} & \begin{tabular}[t]{@{}l@{}} \textit{Real:} [-0.003, 0.003] \\ \textit{Sim:} [-0.1, 0.1] \end{tabular}  & Defines the allowed min/max action. Note: SAC always sees actions in [-1.0, 1.0], but actions are scaled to be within \texttt{sac.action\_range} before executing in the environment. \\
\texttt{sac.discount} & $1 - \frac{1}{H}$ & Discount is defined based on the task horizon $H$ --- see Table~\ref{tab:mdp-def}. \\
\texttt{train.human\_demos} & \begin{tabular}[t]{@{}l|l@{}} \textit{Real:} 3 & \textit{Sim:} 0 \end{tabular} & Human demos collected to seed replay buffer pre-learning \\
\texttt{train.learning\_starts} & \begin{tabular}[t]{@{}l|l@{}} \textit{Real:} 5 & \textit{Sim:} 25 \end{tabular} & Episodes collected via uniformly sampling \texttt{sac. action\_range} to seed the replay buffer pre-learning (this number does not include any human demos) \\
\texttt{train.multi\_grad\_step} & \begin{tabular}[t]{@{}l|l@{}} \textit{Real:} 10 & \textit{Sim:} 20 \end{tabular} & Gradient steps per transition added to the buffer \\
\texttt{train.offline\_multi\_grad\_step} & \begin{tabular}[t]{@{}l|l@{}} \textit{Real:} 0 & \textit{Sim:} 5 \end{tabular} & Gradient steps per transition added to the buffer for episodes collected before \texttt{train. learning\_starts} \\
\texttt{eval.eval\_every} & \begin{tabular}[t]{@{}l|l@{}} \textit{Real:} 30 & \textit{Sim:} 50 \end{tabular} & Number of training episodes between evaluations \\
\texttt{eval.num\_evaluations} & \begin{tabular}[t]{@{}l|l@{}} \textit{Real:} 10 & \textit{Sim:} 64 \end{tabular} & Number of evaluation episodes \\
\bottomrule
\end{tabular}
\end{table*}

\begin{table*}[htbp]
\caption{\textbf{\method Hyperparameters.} \textit{Real} indicates experiments run on the WidowX while \textit{Sim} indicates experiments run on Libero-10 \cite{liu2023libero}.}
\label{tab:sarl_hyperparams}
\centering
\small
\begin{tabular}{llp{5.5cm}}
\toprule
\textbf{Hyperparameter} & \textbf{Value} & \textbf{Description} \\
\midrule
\texttt{outer\_policy.max\_candidates} & \begin{tabular}[t]{@{}l|l@{}} \textit{Real:} 9 & \textit{Sim:} 4  \end{tabular} & Max size of $\cAlang^t$ from Section~\ref{subsec:SARL-algo} and Algorithm~\ref{alg:main}. Max number of language prompts the algorithm may choose from at each step \\
\texttt{outer\_policy.cache\_size} & \begin{tabular}[t]{@{}l@{}} \textit{Real:} See Table~\ref{tab:sarl_task_hyperparams} \\ \textit{Sim:} 32  \end{tabular} &  Max size of $\cAlang$ from Section~\ref{subsec:SARL-algo} and Algorithm~\ref{alg:main}. Total number of unique language prompts the algorithm may learn over \\
\texttt{outer\_policy.action\_dim} & \begin{tabular}[t]{@{}l@{}} \textit{Real:} See Table~\ref{tab:sarl_task_hyperparams}  \\ \textit{Sim:} 32 \end{tabular} & One-hot encoding over cached language commands \\
\texttt{outer\_policy.state\_dim} & 8 & Policy has access to a proprioceptive observation \\
\texttt{outer\_policy.policy\_chunk\_size} & \begin{tabular}[t]{@{}l|l@{}} \textit{Real:} 1 & \textit{Sim:} 10 \end{tabular} & Number of actions used from each evaluation of $\pi$ series policies \cite{intelligence2025pi_} which generate chunks of actions  \\
\texttt{outer\_policy.image\_resolution} & [64, 64] & Image resolution \\
\texttt{outer\_policy.camera\_names} & [``global"] & We only use scene cameras for all experiments \\
\texttt{env.image\_resolution} & [256, 256] & Image resolution of the base environment's camera \\
\texttt{sac.batch\_size} & 256 & Optimization batch size for the replay buffer \\
\texttt{sac.critic\_lr} & 0.0003 & Learning rate for the SAC critic network \\
\texttt{sac.hidden\_dims} & [128, 128, 128] & Hidden dimensions for SAC MLP layers \\
\texttt{sac.cnn\_features} & [32, 32, 32, 32] & Feature maps per layer for the CNN encoder \\
\texttt{sac.cnn\_strides} & [2, 2, 2, 2] & Strides per layer for the CNN encoder \\
\texttt{sac.cnn\_padding} & \texttt{VALID} & Padding type for the CNN encoder layers \\
\texttt{sac.cnn\_latent\_dim} & 50 & CNN output feature dim \\
\texttt{sac.tau} & 0.005 & Target network soft-update rate \\
\texttt{sac.critic\_reduction} & \texttt{min} & Ensemble reduction strategy for target Q-values \\
\texttt{sac.num\_qs} & 10 & Number of Q-networks in the critic ensemble \\
\texttt{sac.dropout\_rate} & 0.0 & Dropout rate applied inside SAC networks \\
\texttt{sac.softmax\_temp} & See Table~\ref{tab:sarl_task_hyperparams} & Softmax temp, $T$, $\pisemt(\ell \mid s) \propto \exp(\Qsem^t(s,\ell)/T)$ from Algorithm~\ref{alg:main}.  \\
\texttt{sac.discount} & $1 - \frac{1}{H}$ & Discount is defined based on the task horizon $H$ --- see Table~\ref{tab:mdp-def}. \\
\texttt{train.human\_demos} & \begin{tabular}[t]{@{}l|l@{}} \textit{Real:} 3 & \textit{Sim:} 0 \end{tabular} & Human demos collected to seed replay buffer pre-learning \\
\texttt{train.learning\_starts} & \begin{tabular}[t]{@{}l|l@{}} \textit{Real:} 5 & \textit{Sim:} 25 \end{tabular} & Episodes collected via uniformly sampling \texttt{sac. action\_range} to seed the replay buffer pre-learning (this number does not include any human demos) \\
\texttt{train.multi\_grad\_step} & \begin{tabular}[t]{@{}l|l@{}} \textit{Real:} 10 & \textit{Sim:} 20 \end{tabular} & Gradient steps per transition added to the buffer \\
\texttt{train.offline\_multi\_grad\_step} & \begin{tabular}[t]{@{}l|l@{}} \textit{Real:} 0 & \textit{Sim:} 5 \end{tabular} & Gradient steps per transition added to the buffer for episodes collected before \texttt{train. learning\_starts} \\
\texttt{eval.eval\_every} & \begin{tabular}[t]{@{}l|l@{}} \textit{Real:} 30 & \textit{Sim:} 50 \end{tabular} & Number of training episodes between evaluations \\
\texttt{eval.num\_evaluations} & \begin{tabular}[t]{@{}l|l@{}} \textit{Real:} 10 & \textit{Sim:} 64 \end{tabular} & Number of evaluation episodes \\
\bottomrule
\end{tabular}
\end{table*}

\begin{table*}[htbp]
\caption{\textbf{\method Task-Specific Hyperparameters}}
\label{tab:sarl_task_hyperparams}
\centering
\small
\begin{tabular}{p{5.5cm}cc}
\toprule
\textbf{Task} & \begin{tabular}[t]{@{}c@{}} \textbf{\texttt{outer\_policy.cache\_size}} \\ (also \textbf{\texttt{outer\_policy.action\_dim}}) \end{tabular} & \textbf{\texttt{outer\_policy.softmax\_temp}} \\
\midrule
\textbf{Real Task 1:} Move the hammer to the plate. Then, grasp the mushroom. & 64 & 0.2 \\
\addlinespace
\textbf{Real Task 2:} Move the banana to the pot on the right and sushi to the bowl. & 100 & 0.5 \\
\addlinespace
\textbf{Real Task 3:} Move the left stuffed toy to the largest bowl. & 32 & 0.1 \\
\addlinespace
\textbf{Real Task 4:} Move the pot lid to the towel. & 32 & 0.2 \\
\addlinespace
\textbf{Libero-10} & 32 & 0.03 \\
\bottomrule
\end{tabular}
\end{table*}

\begin{figure*}[t]
    {\tiny\texttt{\input{prompts/real-gemini-open-ended}}}
    \caption{\footnotesize Prompt for open-ended real-world semantic action candidate generation}
    \label{fig:real-open-ended}
    \vspace{0.3cm}
\end{figure*}

\begin{figure*}[t]
    {\tiny\texttt{\input{prompts/real-gemini-cache}}}
    \caption{\footnotesize Prompt for real-world semantic action candidate generation from cache}
    \label{fig:real-cache}
    \vspace{0.3cm}
\end{figure*}

\begin{figure*}[t]
    {\tiny\texttt{\input{prompts/real-gemini-sp-baseline}}}
    \caption{\footnotesize Prompt for real-world in-context learning VLM baseline}
    \label{fig:real-sp-baseline}
    \vspace{0.3cm}
\end{figure*}

\begin{figure*}[t]
    {\tiny\texttt{\input{prompts/sim-gemini-open-ended}}}
    \caption{\footnotesize Prompt for open-ended sim semantic action candidate generation}
    \label{fig:sim-open-ended}
    \vspace{0.3cm}
\end{figure*}

\begin{figure*}[t]
    {\tiny\texttt{\input{prompts/sim-gemini-cache}}}
    \caption{\footnotesize Prompt for sim semantic action candidate generation from cache}
    \label{fig:sim-cache}
    \vspace{0.3cm}
\end{figure*}

\begin{figure*}[t]
    {\tiny\texttt{\input{prompts/sim-gemini-sp-baseline}}}
    \caption{\footnotesize Prompt for sim in-context learning VLM baseline}
    \label{fig:sim-sp-baseline}
    \vspace{0.3cm}
\end{figure*}

\end{document}